\documentclass[10pt,twocolumn,letterpaper]{article}

\usepackage{cvpr}              

\definecolor{cvprblue}{rgb}{0.21,0.49,0.74}
\usepackage[pagebackref,breaklinks,colorlinks,allcolors=cvprblue]{hyperref}

\def\paperID{*****} 
\def\confName{CVPR}
\def\confYear{2025}

\title{OLATverse: A Large-scale Real-world Object Dataset with Precise Lighting Control}

\author{
Xilong Zhou$^{1}$,
Jianchun Chen$^{*,1}$,
Pramod Rao$^{*,1}$,
Timo Teufel$^{1}$,
Linjie Lyu$^{1}$,\\
Tigran Minasian$^{1}$,
Oleksandr Sotnychenko$^{1}$,
Xiao-Xiao Long$^{2}$,\\
Marc Habermann$^{1}$,
and Christian Theobalt$^{1}$\\[8pt]
$^{1}$Max Planck Institute for Informatics,
$^{2}$Nanjing University\\[4pt]
$^{*}$Equal contribution
}

\usepackage{pifont}
\usepackage{xcolor}
\DeclareUnicodeCharacter{2194}{\ensuremath{\leftrightarrow}}
\DeclareUnicodeCharacter{2195}{\ensuremath{\updownarrow}}

\newcommand{\cmark}{\textcolor{green!70!black}{\ding{51}}} 
\newcommand{\xmark}{\textcolor{red!70!black}{\ding{55}}}             

\definecolor{darkred}{RGB}{200,0,0} 
\definecolor{darkgreen}{RGB}{0,200,0} 

\newcommand{\refFig}[1]{Fig.~\ref{fig:#1}}
\newcommand{\refTab}[1]{Tab.~\ref{tab:#1}}
\newcommand{\refSec}[1]{Sec.~\ref{sec:#1}}
\newcommand{\refEq}[1]{Eq.~\ref{eq:#1}}

\newcommand{\ourdata}{\textbf{OLATverse}}

\newcommand{\bgimg}{\mathbf{I}_\text{bg}}
\newcommand{\fgimg}{\mathbf{I}_\text{fg}}
\newcommand{\bgmat}{\mathbf{M}_1}
\newcommand{\rmbg}{\mathbf{M}_2}
\newcommand{\sam}{\mathbf{M}_3}
\newcommand{\maskstup}{\mathbf{M}_\text{stup}}
\newcommand{\maskobj}{\mathbf{M}_\text{obj}}

\newcommand{\Ip}{\mathbf{I}^+}
\newcommand{\In}{\mathbf{I}^-}

\newcommand{\Icg}{\mathbf{I}_\text{cg}}
\newcommand{\Icgp}{\mathbf{I}_\text{cg}^+}
\newcommand{\Icgn}{\mathbf{I}_\text{cg}^-}
\newcommand{\Ic}{\mathbf{I}_{\perp}}
\newcommand{\Icp}{\mathbf{I}_{\perp}^+}
\newcommand{\Icn}{\mathbf{I}_{\perp}^-}

\newcommand{\normal}{\mathbf{N}}

\newcommand{\diffuse}{\mathbf{D}}

\newcommand{\maskolat}{\mathbf{M}_\text{i}}
\newcommand{\tgtenv}{\mathbf{E}}
\newcommand{\olat}{\mathbf{I}_\text{i}}
\newcommand{\relit}{\mathbf{I}_\text{relit}}
\newcommand{\aver}{\mathcal{F}}

\newcommand{\imgnumber}{9M}
\newcommand{\lvisnumber}{18.5}
\newcommand{\objnumber}{765}

\usepackage{multirow}

\definecolor{junglegreen}{rgb}{0.113, 0.639, 0.5}

\definecolor{ferrarired}{rgb}{1, 0.157, 0}

\begin{document}

\maketitle
\begin{abstract}

We introduce \ourdata{}, a large-scale dataset comprising around $\imgnumber$ images of $\objnumber$ real-world objects, captured from multiple viewpoints under a diverse set of precisely controlled lighting conditions. While recent advances in object-centric inverse rendering, novel view synthesis and relighting have shown promising results, most techniques still heavily rely on the synthetic datasets for training and small-scale real-world datasets for benchmarking, which limits their realism and generalization. To address this gap, \ourdata{} offers two key advantages over existing datasets: large-scale coverage of real objects and high-fidelity appearance under precisely controlled illuminations. Specifically, \ourdata{} contains $\objnumber$ common and uncommon real-world objects, spanning a wide range of material categories. 
Each object is captured using $35$ DSLR cameras and $331$ individually controlled light sources, enabling the simulation of diverse illumination conditions.
In addition, for each object, we provide well-calibrated camera parameters, accurate object masks, photometric surface normals, and diffuse albedo as auxiliary resources. 
We also construct an extensive evaluation set, establishing the first comprehensive real-world object-centric benchmark for inverse rendering and normal estimation. 
We believe that \ourdata{} represents a pivotal step toward integrating the next generation of inverse rendering and relighting methods with real-world data.
The full dataset, along with all post-processing workflows, will be publicly released at \url{https://vcai.mpi-inf.mpg.de/projects/OLATverse/}.

\end{abstract}    
\section{Introduction}
\label{sec:intro}











The appearance of real-world objects is the result of the complex interaction between geometry, material, and lighting conditions. Acquiring high-fidelity material and geometry of objects, and synthesizing photorealistic appearance under novel illuminations, remain fundamental challenges in computer vision and computer graphics, with widespread applications in film and gaming industries, autonomous driving, robotics, and VR/AR. With the advances in deep learning and generative models, recent years have witnessed rapid progress in 3D generation and reconstruction~\cite{shimvdream,toussaint2025probesdf, wang2021neus, wang2023neus2, long2024wonder3d,poole2022dreamfusion, siddiqui2024meta}, relighting~\cite{jin2024neural,zeng2024dilightnet, zhang2025scaling, zeng2023relighting, liu2024bigs} and inverse rendering~\cite{bi2024gs3,zhang2021nerfactor,liang2024gs,gao2024relightable,zeng2024rgb,liang2025diffusion}. However, due to the lack of high-fidelity large-scale real-world datasets, the majority of these techniques are trained on synthetic datasets~\cite{deitke2023objaverse, wu2023omniobject3d} or evaluated on small-scale real datasets~\cite{bi2024gs3, liu2023openillumination}. As a result, the synthesized results are typically limited in the realism, and their performance in real-world scenarios cannot be reliably assessed because of the significant domain gap between real and synthetic data.

\begin{figure*}[t]
    \centering
    \includegraphics[width=\linewidth]{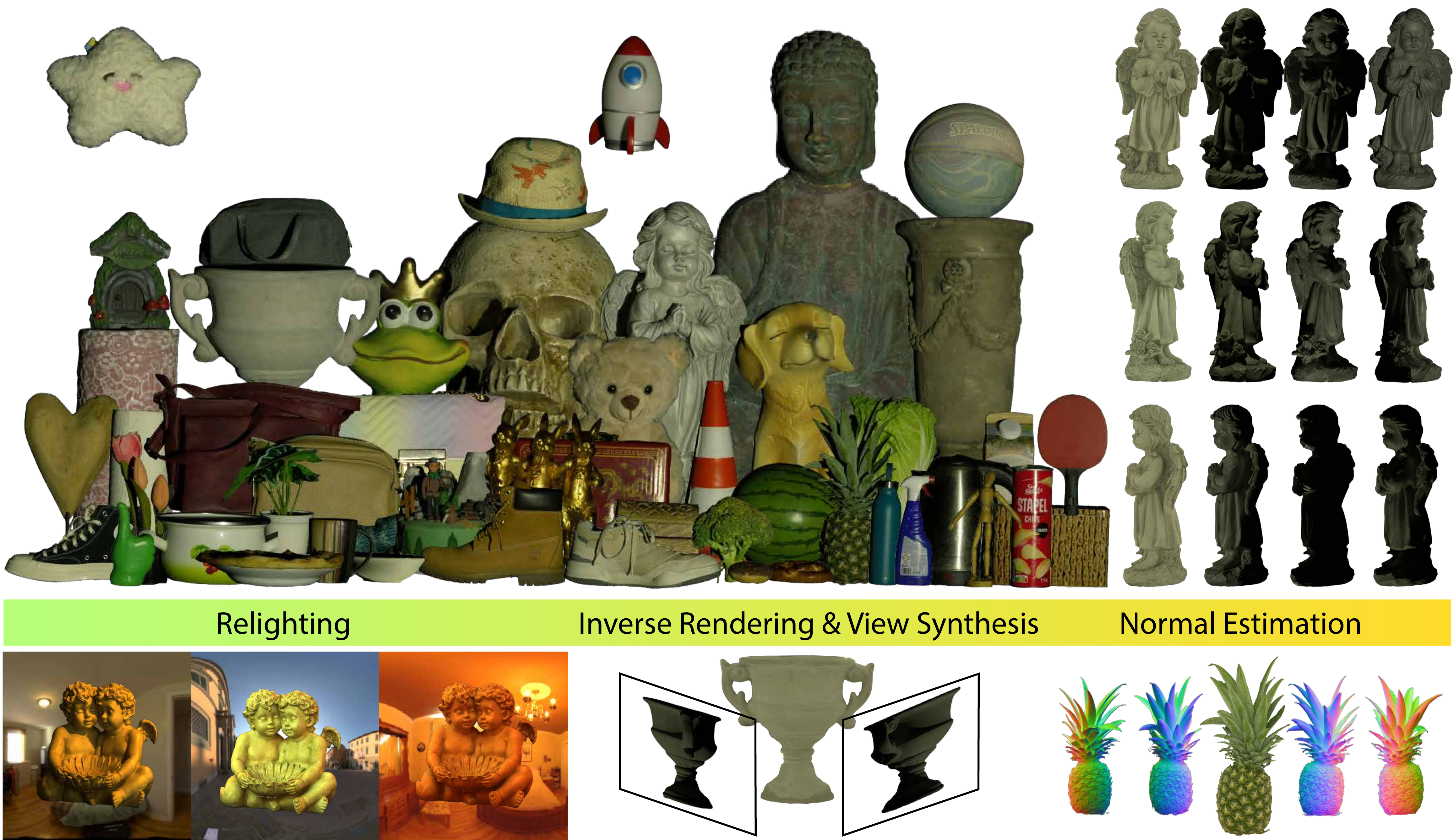} 
    \vspace{-0.2in}
    \caption{\ourdata{} is a large-scale high-quality real-world OLATs dataset comprising around $\imgnumber$ images of $\objnumber$ objects. We illustrate a subset of objects (top left), demonstrating the diversity and large-scale coverage of \ourdata{}, and we also show one sample object captured under multiple viewpoints and precisely controlled single light sources (top right). \ourdata{} can be utilized as a comprehensive real-world benchmark for inverse rendering, novel view synthesis and normal estimation.}
    \label{fig:teaser}
    \vspace{-0.1in}
\end{figure*}

Considerable effort has recently been devoted to constructing large-scale and high-fidelity object-centric datasets. However, due to complexity in hardware and data processing pipelines, the existing datasets remain constrained in at least one of three key aspects: quality~\cite{shi2017learning, collins2022abo, deitke2023objaverse}, scale~\cite{jensen2014large, zhou2013multi,kuang2022neroic,grosse2009ground,toschi2023relight,zeng2023relighting, kuang2023stanford, bi2024gs3, dihlmann2024subsurface,liu2023openillumination} or precise lighting control~\cite{wu2023omniobject3d,dong2025digital}. As summarized in \refTab{dataset}, the first category of datasets~\cite{shi2017learning, collins2022abo, deitke2023objaverse} comprises synthetic and hybrid objects. While the scale of these datasets is large, they typically lack realism and exhibit significant variance in object quality. The second category of datasets~\cite{wu2023omniobject3d, dong2025digital} contains thousand-scale real-world objects, offering both realism and scale. However, the material creation in these datasets heavily relies on manual annotation, and the limited illumination setups prevent accurate simulation of complex light transport. The third group of datasets~\cite{bi2024gs3,liu2023openillumination,dihlmann2024subsurface} captures objects under precisely controlled illumination, but this group is typically small in both scale and diversity, restricting their applications in comprehensive benchmarking and generative models training. To the best of our knowledge, no existing object dataset simultaneously provides both large-scale coverage and high-fidelity appearance.

To address these challenges, we propose \ourdata{}, the first large-scale real dataset that provides high-fidelity images of a diverse set of objects captured under precisely-controlled illumination and camera configurations. 
Our dataset consists of $\objnumber$ real-world objects, covering a wide range of material categories (e.g., wood, stone, leather, plastic, metal, paper, plaster, fabric, ceramic) and $\lvisnumber \%$ 
LVIS categories~\cite{gupta2019lvis} from common to uncommon objects. \ourdata{} is acquired using lightstage setup ~\cite{debevec2000acquiring}, which enables One-Light-at-a-Time (OLAT) capture of real-world objects and provides rich information for analyzing the complex interaction between surface reflectance properties and light sources. Specifically, each object is captured using 35 well-calibrated DSLR cameras and illuminated by 331 individually controlled light sources, simulating a set of illuminations including uniform lightings, OLATs, gradient illumination and pre-defined environmental illuminations. In total, this setup yields over $\imgnumber$ high-fidelity images. In addition, we develop efficient semi-automatic mask processing pipeline to extract high-quality masks for each object captured under multiple views. Furthermore, we employ polarized gradient illumination to recover surface normal and diffuse albedo, following techniques proposed by Ma et al.~\cite{ma2007rapid}. 
These auxiliary data are particularly valuable for evaluating and supervising multi-modal tasks. 

\begin{table*}[ht]
\centering
\scriptsize
\caption{Comparison of object-centric datasets targeting inverse rendering and relighting tasks. We list a detailed comparison of \ourdata{} with existing datasets across several key attributes. The compared aspects include number of object (\textbf{$\#$ Objs}), whether data source is real (\textbf{Real}), lighting conditions (\textbf{IllumCond}), number of illuminations (\textbf{$\#$ Illum}), number of views (\textbf{$\#$ Views}), and capture device (\textbf{Device}).  In the column of \textbf{IllumCond}, ENV denotes environment illumination, PAT represents pattern illumination. \textit{Unspec.} indicates that the corresponding information is not specified in the dataset. $()$ indicates that only a small portion of the dataset satisfies the criterion.}
\vspace{-0.1in}
\resizebox{\textwidth}{!}
{
\begin{tabular}{lcccccccr}
\hline\hline
\textbf{Dataset}  & \textbf{$\#$ Objs} & \textbf{Real}  & \textbf{IllumCond} & \textbf{$\#$ Illum}   & \textbf{$\#$ Views} & \textbf{Device} \\

\hline

ABO~\cite{collins2022abo} & 8K & \xmark  & ENV & 3 & -- & -- \\

ShapeNet-Intr~\cite{shi2017learning}  & 31K & \xmark  & ENV & 36 & -- & -- \\

TexVerse~\cite{zhang2025texverse} & 818K & (\cmark)   & -- & -- & -- & -- \\

Objaverse~\cite{deitke2023objaverse} & 858K & (\cmark)    & -- & -- & -- & --\\

\hline


NeROIC~\cite{kuang2022neroic} & 3 & \cmark & ENV & 4$\sim$6 & 40 & camera \\

Standford-ORB~\cite{kuang2023stanford}  & 14 & \cmark & ENV & 7 & 70 & scanner+camera \\

DTC~\cite{dong2025digital} & 2K & \cmark  & (ENV) & 2 & 120  & scanner+camera \\

OmniObject3D~\cite{wu2023omniobject3d} & 6K & \cmark   & -- & -- & -- & scanner \\

\hline

DiLiGenT-MV~\cite{zhou2013multi} & 5 & \cmark & OLAT & 96  & 20 & scanner \\

GS$^3$~\cite{bi2024gs3}  & 6 & \cmark    & OLAT & \textit{Unspec.}  & \textit{Unspec.} & lightstage \\

NRHint~\cite{zeng2023relighting}& 7  & \cmark   & OLAT & \textit{Unspec.}  & \textit{Unspec.}  & smartphones  \\

DIR~\cite{choi2025real} & 16 & \cmark  & OLAT & 144 & 2  & LCD display \\

MIT-Intrinsic~\cite{grosse2009ground} & 20 & \cmark  & OLAT & 10  & 1 & camera\\

ReNe~\cite{toschi2023relight} & 20 & \cmark    & OLAT & 40 & 50 & robots\\

SSS-GS~\cite{dihlmann2024subsurface} & 20 & \cmark  & OLAT & 167 & 158 & lightstage \\

OpenIllumination~\cite{liu2023openillumination}  & 64 & \cmark  & PAT+OLAT & 13+142 & 72 & lightstage\\

OpenSubstance~\cite{pei2025opensubstance}  & 187 & \cmark  & PAT+OLAT & 16+1620 & 270 & lightstage\\

\textbf{Ours} & \textbf{\objnumber} & \cmark & \textbf{ENV+OLAT} & \textbf{11+331} & \textbf{35} & \textbf{lightstage}\\

\hline\hline
\end{tabular}

}
\vspace{-0.2in}
\label{tab:dataset}
\end{table*}

We showcase the applications of \ourdata{} across multiple tasks. The linearity of light transport allows recomposing the captured OLAT images to synthesize object appearance lit under any arbitrary novel illuminations, enabling the creation of large-scale training data for generative priors. Furthermore, we curate a subset of 42 objects with diverse material categories to construct a comprehensive evaluation benchmark. Using this benchmark, we conduct representative baseline experiments on inverse rendering, relighting, view synthesis, and normal estimation. These applications demonstrate the potential of \ourdata{} to advance future research in realistic 3D vision and relighting, facilitating the integration of data-driven methods with real-world datasets in the graphics and vision communities. 

In summary, our contributions are as follows:

\begin{itemize}
    \item We introduce \ourdata{}, the first publicly available large-scale real-world dataset, comprising around $\imgnumber$ images of $\objnumber$ objects with diverse material and object categories captured under precise lighting control. 
    \item Each object is captured under precisely controlled lighting conditions, including uniform illumination, OLATs, environmental illumination, and gradient illumination.
    \item We provide auxiliary data, consisting of well-calibrated camera parameters, accurate object masks, diffuse albedo, and surface normals recovered via polarized gradient illumination.
    \item We establish \ourdata{} as a comprehensive real-world benchmark for multiple tasks, and highlight its potential as a valuable resource for generative prior learning.
\end{itemize}

\section{Related Work}
\label{sec:related}

\subsection{Object-centric Datasets}

The acquisition of high-fidelity and large-scale datasets is critical for advancing data-driven methods in inverse rendering and relighting tasks. Recent research has introduced different datasets that capture the geometry and reflectance properties of full-body human avatars~\cite{zhou2023relightable, stratou2011effect,wang2025relightable, teufel2025humanolat}, human faces~\cite{saito2024relightable,zhang2021neural,prao20253dpr}, planar material surfaces~\cite{deschaintre2018single, vecchio2024matsynth, ma2023opensvbrdf, zhou2023photomat, zhou2021adversarial} and 3D objects~\cite{shi2017learning,collins2022abo,deitke2023objaverse,jensen2014large,zhou2013multi,kuang2022neroic,grosse2009ground,toschi2023relight,zeng2023relighting,wu2023omniobject3d,kuang2023stanford,dong2025digital,bi2024gs3,dihlmann2024subsurface,liu2023openillumination,zhang2025texverse, choi2025real}. In comparison, our work focuses on appearance capture of real-world objects. We categorize the existing object-centric datasets into several subgroups, as summarized in \refTab{dataset}. The first group focuses on synthetic objects, which are highly scalable due to the procedural generation pipelines and the availability of public 3D models. ShapeNet-Intrinsics~\cite{shi2017learning}, for example, constructs a large-scale object intrinsic dataset from 3D models in the ShapeNet~\cite{chang2015shapenet} collection. ABO~\cite{collins2022abo} introduces synthetic household objects with complex geometry and materials. Objaverse~\cite{deitke2023objaverse} and TexVerse~\cite{zhang2025texverse} significantly expand the diversity of objects by incorporating both synthetic and scanned real assets. While these datasets are large-scale, the quality of these datasets varies significantly across different objects, and a domain gap still remains between synthetic and real-world data. Consequently, effective curation strategies are required to filter out low-quality data, limiting their applicability in realistic 3D vision and relighting.

\begin{figure*}[t]
    \centering
    \includegraphics[width=\linewidth]{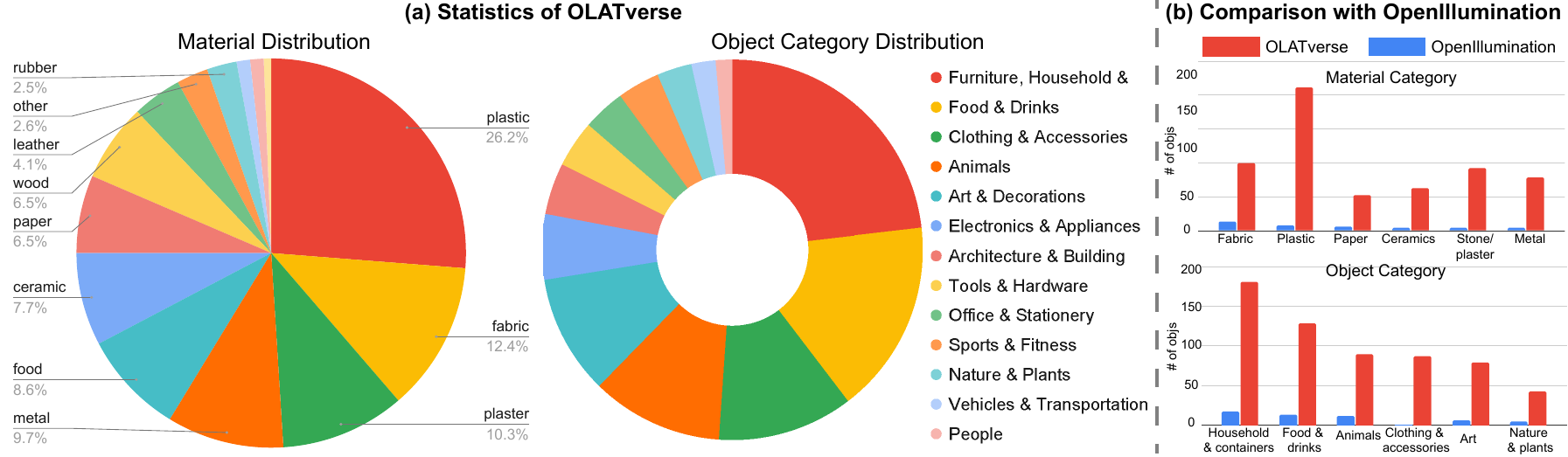}     \vspace{-0.2in}
    \caption{ (a) We visualize the statistics of $\ourdata{}$, including the material distribution and high-level object category distribution. (b) We also show comparison against OpenIllumination~\cite{liu2023openillumination} for the six largest material and object categories in terms of object count.}
    \label{fig:dataset_overview}
    \vspace{-0.1in}
\end{figure*}

The second group of datasets focuses on collecting real-world objects with high-fidelity appearance. Such datasets are typically collected either in specialized studios or in-the-wild environments. Among this group, DTC~\cite{dong2025digital} and OmniObject3D~\cite{wu2023omniobject3d} provide the meshes of thousands of real objects via 3D scanning. Although these two datasets are large-scale, they offer limited support for appearance analysis and lack control over lighting conditions. Specifically, OmniObject3D does not support appearance capture, and DTC only contains only 50 objects captured under two environmental illuminations. In addition, the material maps provided in DTC are manually created by artists to approximate the realistic appearance, which introduces a gap to the real-world appearance. 

The third group of datasets~\cite{bi2024gs3,liu2023openillumination,dihlmann2024subsurface} captures OLATs in studios with a well-designed lighting setups (e.g., lightstage~\cite{debevec2000acquiring}, LCD, smartphone flash, etc.), facilitating detailed analysis of light transport. However, due to the complexity of the setup and capture procedure, these datasets remain small in scale and limited in diversity. The largest OLAT dataset, OpenIllumination~\cite{liu2023openillumination}, contains only $64$ objects, limiting its application for comprehensive benchmarking and generative model training. In comparison with existing object-centric datasets, \ourdata{} encompasses $\objnumber$ real-world objects spanning diverse material categories and provides high-fidelity appearance under precisely controlled lighting, thereby achieving a combination of scale, diversity, and realism.

\subsection{Inverse Rendering $\&$ Relighting }

The goal of inverse rendering is to recover the  intrinsic properties of scenes or objects (e.g., albedo, normal, roughness, and metallicity) to reproduce plausible renderings under novel illuminations. Traditional approaches typically leverage advanced differentiable renderers~\cite{nimier2019mitsuba, ravi2020accelerating,Laine2020diffrast} under reliable constraints provided by geometry to jointly optimize intrinsic properties and lighting by minimizing the discrepancy between input and rendered images. With the advances in deep learning, Neural Radiance Fields (NeRF)\cite{mildenhall2021nerf} is introduced to jointly encode geometry and appearance in a single multi-layer perceptron (MLP), producing photorealistic rendering via volume rendering. NeRF demonstrates superior performance but suffers from slow rendering speed. To address this limitation, 3D Gaussian Splatting (3DGS)\cite{kerbl20233d} is proposed to represent scenes with localized Gaussian kernels and can offer real-time efficiency through an efficient differentiable splatting process. Leveraging the benefits of these two representations, many recent inverse rendering methods~\cite{zhang2021nerfactor,shi2025gir, gao2024relightable, liang2024gs, du2024gs, zhang2024prtgaussian,liu2024bigs, jin2023tensoir, fan2025rng} extend NeRF or 3DGS with additional optimizable Bidirectional Reflectance Distribution Function (BRDF) modules and a differentiable renderer to recover intrinsic properties and lighting from multi-view input images. More recently, with the rapid progress of diffusion-based generative models, a new branch of research leverages the strong priors of pretrained diffusion models for inverse rendering~\cite{zeng2024rgb,ye2024stablenormal, liang2025diffusion, chen2024intrinsicanything} and relighting~\cite{jin2024neural, zeng2024dilightnet} tasks. These methods typically fine-tune pretrained diffusion models on datasets annotated with material maps or renderings under novel illuminations. During inference, the diffusion model extracts material reflectance properties or relit images through inverse denoising steps. In this work, we demonstrate that \ourdata{} serves as a comprehensive benchmark for object-centric inverse rendering and relighting tasks, and we evaluate several representative approaches as baselines on our selected validation dataset.





\section{OLATverse}
\label{sec:realolat}

In this section, we first provide an overview of \ourdata{}, introducing the object and illumination statistics (\refSec{overview}). We then describe the hardware configuration (\refSec{setup}), and the data post-processing pipeline (\refSec{processing}), including camera and light calibration, mask segmentation, photometric surface normal estimation, and diffuse albedo extraction.

\subsection{Overview}
\label{sec:overview}

\begin{figure*}
    \centering
    \includegraphics[width=\linewidth]{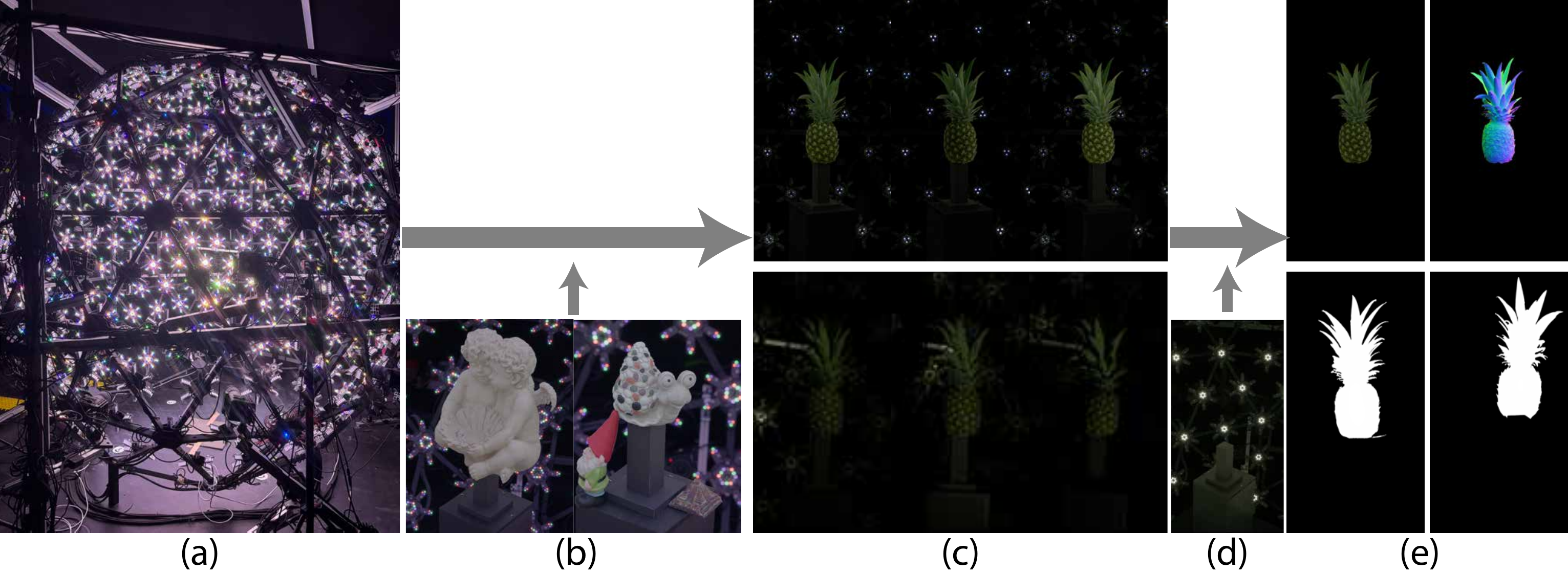} 
    \vspace{-0.3in}
    \caption{ Illustration of the dataset capture setup and process pipeline. We utilize wooden stands with varying sizes and (a) a lightstage setup to capture raw videos of objects. During the calibration session, we record (b) reference objects to extract accurate camera parameters, which are utilized to extract (c) undistorted OLALs and relit images under varying illuminations from raw videos. Next, we capture (d) background stand image and perform (e) semi-automatic mask segmentation and normal extraction for each object. }
    \label{fig:pipeline}
    \vspace{-0.2in}
\end{figure*}

\paragraph{Object Composition} \ourdata{} comprises $\objnumber$ real-world objects with high diversity in material types, object categories, and physical sizes. As shown in (a) of \refFig{dataset_overview}, it covers over $13$ material categories, including wood, stone, plaster, fabric, plastic, metal, food, plant, ceramics, leather, wax, rubber, and paper. Furthermore, \ourdata{} consists of a wide range of common and uncommon real-world objects, spanning over $\lvisnumber\%$ of LVIS categories~\cite{gupta2019lvis}, substantially outperforming existing real-world object datasets, such as OmniObject3D ($10.8\%$)~\cite{wu2023omniobject3d}, OpenIllumination ($4\sim5\%$)~\cite{liu2023openillumination} and DTC ($3\%$)~\cite{dong2025digital}. Additionally, \ourdata{} is not limited to object physical dimension, encompassing a wide range of sizes from $5\mathrm{cm}$ to $100\mathrm{cm}$, which is broader than OpenIllumination dataset ($10 \sim 20 \mathrm{cm}$). These attributes highlight the diversity and scale of \ourdata{}, establishing it as the first comprehensive OLAT dataset of real-world objects.

\begin{figure}[t]
    \centering
    \includegraphics[width=\linewidth]{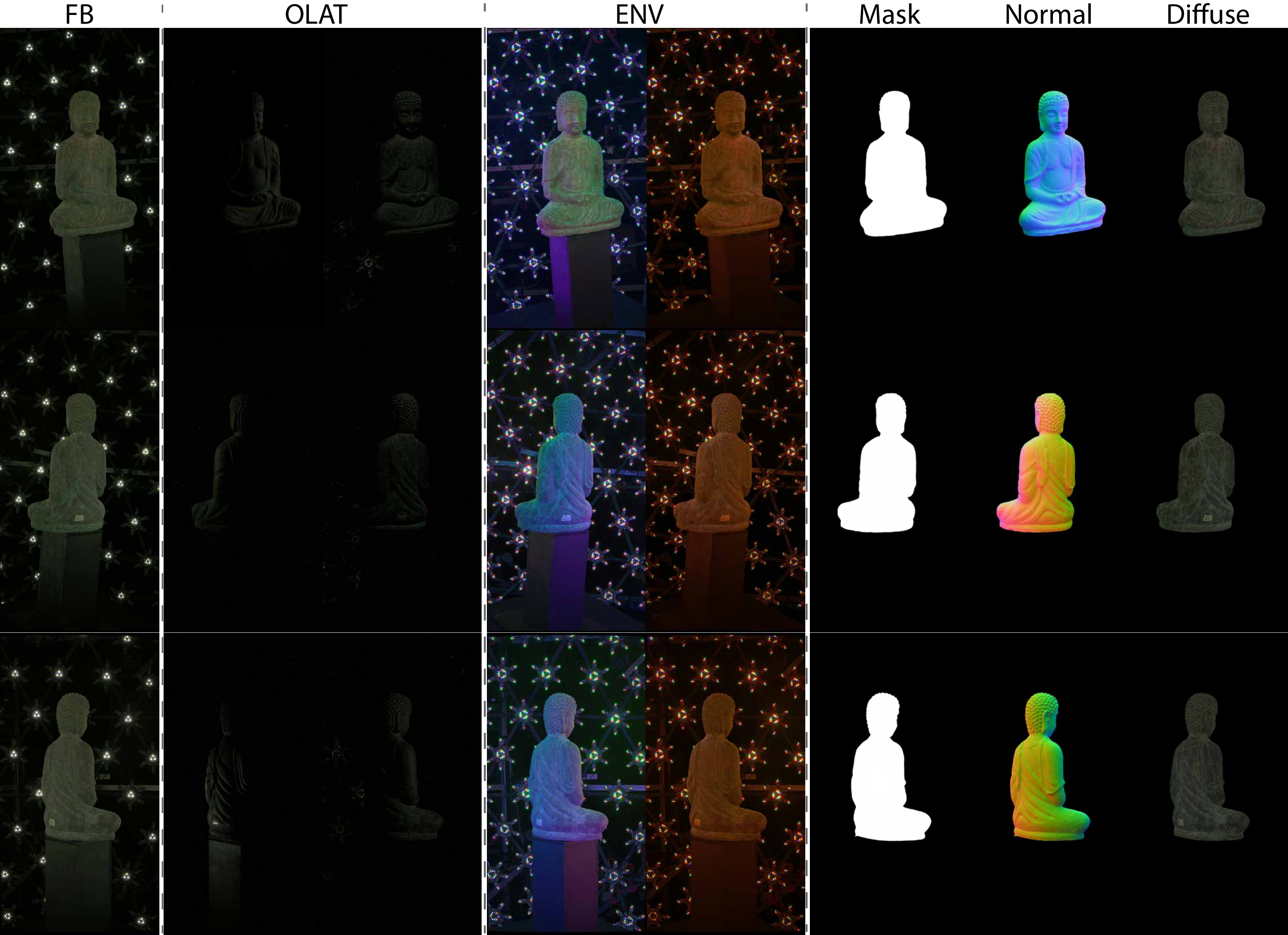}
    \vspace{-0.2in}
    \caption{We visualize one sample of \ourdata{}, which includes full bright (FB), OLATs, relit images under varying pre-defined environmental illuminations (ENV), object mask, surface normals, and diffuse albedo.}
    \label{fig:data_example}
    \vspace{-0.2in}

\end{figure}

\paragraph{Illumination Composition} In \ourdata{}, each object is captured using $35$ DSLR cameras positioned at different viewpoints under a diverse set of lighting conditions, including $1$ uniform white illumination, $12$ polarized gradient illumination, $10$ environment illuminations, and $331$ OLATs, resulting in approximately $12\mathrm{K}$ high-quality images per object. Specifically, uniform white illumination are utilized for mask segmentation and mesh reconstruction; gradient illumination are utilized in surface normals and diffuse albedo extraction; environment illuminations and OLATs capture the high-fidelity reflectance properties under precisely-controlled lighting and enable image-based rendering under novel illuminations. Some representative samples in \ourdata{} are visualized in \refFig{teaser} and ~\refFig{data_example}. 

\paragraph{Comparisons} We compare \ourdata{} against one representative OLAT real dataset, OpenIllumination~\cite{liu2023openillumination}. We visualize the number of objects in the six largest material and high-level object categories, as illustrated in \refFig{dataset_overview} (b). While OpenIllumination covers the similar material and object categories as \ourdata{}, the scale of each category is significantly smaller than our dataset. In addition, OpenIllumination does not provide auxiliary resources such as surface normals and diffuse albedo, limiting its applications in the multi-modal tasks.

\subsection{Capture Setup}
\label{sec:setup}

We capture \ourdata{} using lightstage, as shown in (a) of \refFig{pipeline}, a spherical dome equipped with 35 RED Komodo 6K cameras and 331 controllable LEDs capable of emitting red, green, blue, amber, and white light (RGBAW). The cameras and lights are distributed $360^\circ$ around the center of the dome, enabling synchronized multi-illumination capture at 30 FPS. 
Consistently capturing large real-world objects of varying sizes requires a carefully designed setup.
At the center of the lightstage, we place a table and wooden solid stands with different sizes to support the objects of varying physical sizes. 
We select the stand with a surface area slightly smaller than the objects to minimize occlusion from the lighting located on the lower lightstage dome. Additionally, the table and stands are all covered with a black blanket or dark matte paper, to avoid unnecessary artifacts introduced by color bleeding and specular reflections from the supports. Furthermore, we manually adjust the focus of camera lenses to maintain consistent image quality, ensuring that objects of varying sizes appear properly scaled within the capture frame.

\subsection{Data Processing}
\label{sec:processing}

\begin{figure}[t]
    \centering
    \includegraphics[width=\linewidth]{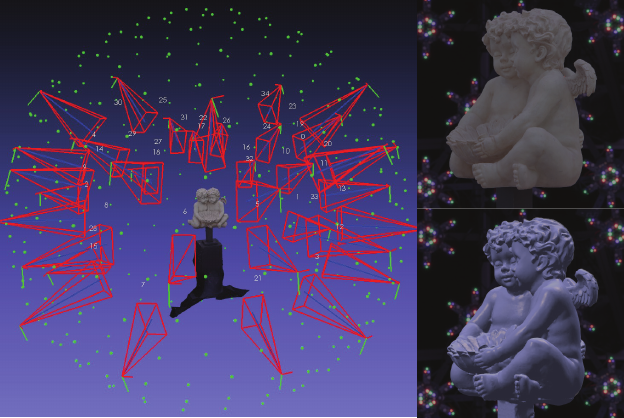}     
    \vspace{-0.2in}
    \caption{Visualization of calibrated cameras (marked as \textcolor{darkred}{red}), light sources (marked as \textcolor{darkgreen}{green}) and object mesh in our capture setup. In the left part of the figure, the reconstructed mesh decently matches the original image, demonstrating the correctness of the calibration process. }
    \label{fig:calibration}
    \vspace{-0.2in}
\end{figure}

\paragraph{Camera and Lighting Calibration}
Acquiring accurate camera parameters consistently across large-scale $\objnumber$ objects in a studio setup is a non-trivial task. Directly applying feature-based calibration algorithms to different objects leads to unstable and ambiguous calibration quality since the real-world objects vary widely in physical dimension, texture richness, and material properties, which limit the accuracy and consistency of the camera calibration process. To address this challenge, we leverage the fixed camera configuration of the lightstage, and perform a calibration session every $20\sim30$ regular capture sessions, demonstrated in (b) of \refFig{pipeline}. During calibration session, we lock down several fixed objects with rich texture and Lambertian surfaces as calibration references, and employ the feature-based algorithms implemented in Metashape~\cite{agisoft_metashape} to recover both intrinsic and extrinsic camera parameters. For subsequent capture sessions, the previously estimated camera parameters are reused. All calibration captures are conducted under uniform white illumination to ensure robust and consistent feature detection. We quantitatively evaluate the calibration accuracy by computing the mean reprojection error of triangulated keypoints across the reference objects, yielding an average error of $0.86$ pixels. The positions of individual lights are also provided within the same canonical system by measuring their physical location. We visualize the calibrated cameras, light sources, and the reconstructed mesh of one calibration reference object in ~\refFig{calibration}. 
As is shown, the reconstructed mesh qualitatively aligns with real captured frames, demonstrating the accuracy of calibration process.

\paragraph{Mask Segmentation}

Unlike human-centric datasets~\cite{teufel2025humanolat}, where Sapiens~\cite{kim2025sapiensid} is tailored specifically for human body segmentation, efficiently producing high-quality masks for large-scale objects in our capture setup remains a challenging task. A straightforward solution is to employ Segment-Anything (SAM)~\cite{kirillov2023segment} with multiple bounding-box prompts for instance segmentation, similar to OpenIllumination~\cite{liu2023openillumination}. However, this strategy is inefficient and difficult to scale for a large-scale dataset. To efficiently tackle this challenge, we develop a simple yet effective semi-automatic segmentation pipeline that combines background matting (bgMatting)~\cite{lin2021real}, SAM~\cite{kirillov2023segment} and RMBG-2.0~\cite{bria2023rmbg2}. Specifically, we capture a foreground image $\fgimg$, containing the object and stand, and a background image $\bgimg$ only for the stand, as is shown in (d) of \refFig{pipeline}. We empirically observe that bgMatting is capable of roughly separating the object from the stand, but fails to preserve fine details around the contour of objects. SAM, when guided by bounding boxes generated from bgMatting, can achieve a clean separation between the stand and the object but at the cost of mask quality. In comparison to the other two methods, RMBG-2.0 captures the most detailed contours but consistently treats the stand as part of the foreground, yielding a wrong object mask. Therefore, we leverage the strengths of all three segmentation strategies, and compute the final object mask $\maskobj$ as follows:

\begin{figure}[t]
    \centering
    \includegraphics[width=\linewidth]{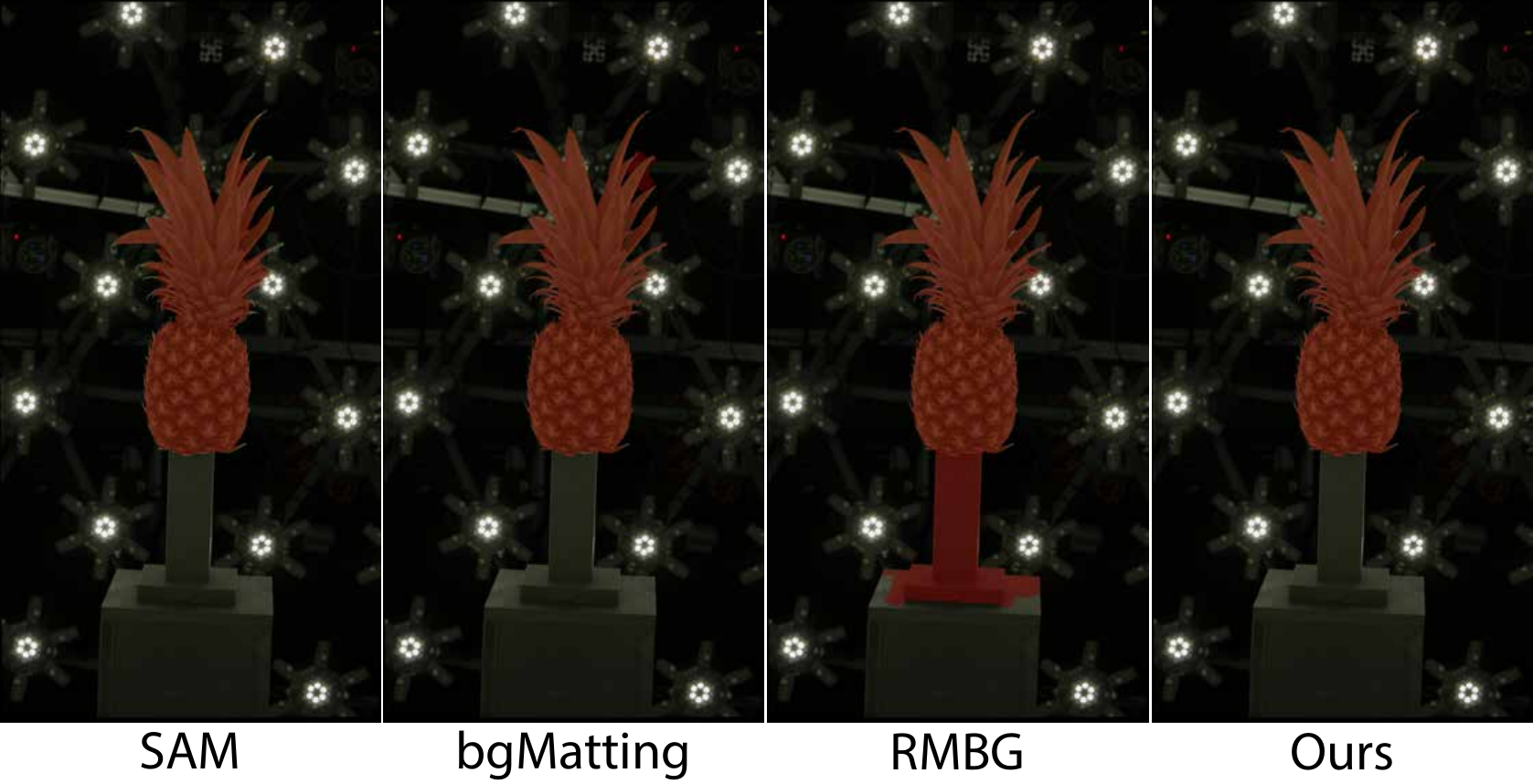} 
    \vspace{-0.3in}
    \caption{Semi-automatic mask processing. We show the object masks generated by SAM, bgMatting, RMGB-2.0 as well as our final mask produced by our proposed mask segmentation strategy. }
    \label{fig:semi-mask}
    \vspace{-0.2in}
\end{figure}

\begin{figure*}[t]
    \centering
    \includegraphics[width=\linewidth]{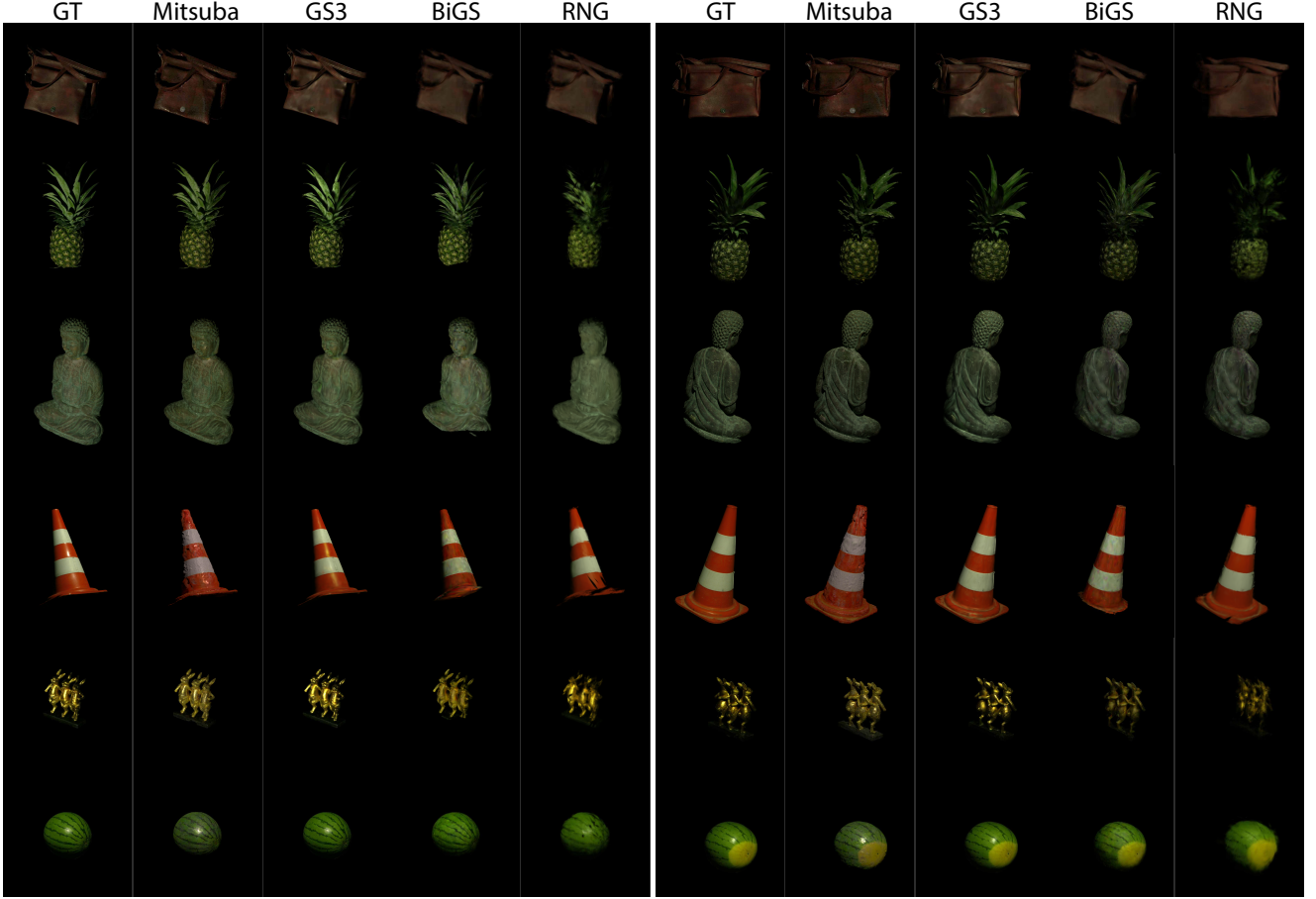}    
    \vspace{-0.2in}
    \caption{We visualize the inverse rendering and novel view synthesis results of several baseline methods (Mitsuba~\cite{nimier2019mitsuba}, GS$^3$~\cite{bi2024gs3},BiGS~\cite{liu2024bigs} and RNG~\cite{fan2025rng}) evaluated on our validation dataset. In this figure, we show relit objects from inference views and light directions.}
    \label{fig:IR_img}
    \vspace{-0.2in}
\end{figure*}

\begin{equation}
\maskstup = \left\{
\begin{array}{ll}
\rmbg(\bgimg) & \text{(a)}, \\
\rmbg(\bgimg) [1 - \sam(\bgmat(\bgimg, \fgimg))] & \text{(b)},
\end{array}
\right.
\label{eq:mask_1}
\end{equation}

\begin{equation}
\maskobj^* = \rmbg(\fgimg)(1 - \maskstup),
\end{equation}

where $\bgmat$, $\rmbg$ and $\sam$ represent bgMatting~\cite{lin2021real}, SAM~\cite{kirillov2023segment} and RMBG-2.0~\cite{bria2023rmbg2}, respectively. $\maskstup$ represents the mask of the stand. In \refEq{mask_1}, \textit{case a} is applied to lower camera views, where part of the objects may be occluded by the stand, while \textit{case b} is used for all other views. Then we refine the intermediate results $\maskobj^*$ using morphological transformations and remove the disconnected components to obtain the final clean mask $\maskobj$. This pipeline achieves a success rate of $95\%$ across all objects from all views in our setup. For the remaining failure cases, we create a lightweight user interface for manual correction. In ~\refFig{semi-mask}, we illustrate masks generated by three individual segmentation strategies and our method.

\paragraph{Normal and Albedo Extraction} In addition to diverse illuminations, \ourdata{} provides pseudo ground truth surface normals and diffuse albedo as auxiliary data to support multimodal tasks. We recover surface normals through photometric stereo solutions~\cite{guo2019relightables} by analyzing radiance variations of images captured under gradient illuminations. 
For all cameras, we captured images $\Icg$ under color gradient illumination. 
To further cancel out the view-dependent specular reflection on the non-Lambertian surfaces~\cite{ma2007rapid, ghosh2009estimating}, we apply linear polarized filters to five fixed cameras, and capture images under polarized gradient full bright illuminations, denoted as $\Ic$. The diffuse albedo $\diffuse$ and surface normals $\normal$ are computed as follows:

\begin{equation}
\diffuse = 0.5 (\Icp + \Icn),
\label{eq:diff}
\end{equation}

\vspace{-0.1in}

\begin{equation}
 \normal^* = \frac{(\Ip - \In)}{(\Ip + \In)}, \quad \normal = \frac{\normal^*}{| \normal^* |},
\label{eq:normal}
\end{equation}

where $\Ip$ and $\In$ denote image captured under opposite gradient directions in either polarized or non-polarized settings. For polarized normals, we utilize polarized captures $\Icp$ and $\Icn$, whereas for non-polarized normals, the images under color gradient illuminations $\Icgp$ and $\Icgn$ are used. Both polarized and non-polarized normals are incorporated in the final dataset, and we empirically observe that polarized normals yield higher accuracy for most objects. 

\section{Application}
\label{sec:application}

In this section, we demonstrate several applications of \ourdata{}. First, we exploit the linearity of light transport to synthesize relit images under arbitrary novel illuminations using OLATs (\refSec{relighting}). Then we set up a validation dataset and conduct baseline experiments on inverse rendering and view synthesis (\refSec{irnvs}), as well as surface normal estimation (\refSec{normal}).

\begin{figure*}[t]
    \centering
    \includegraphics[width=\linewidth]{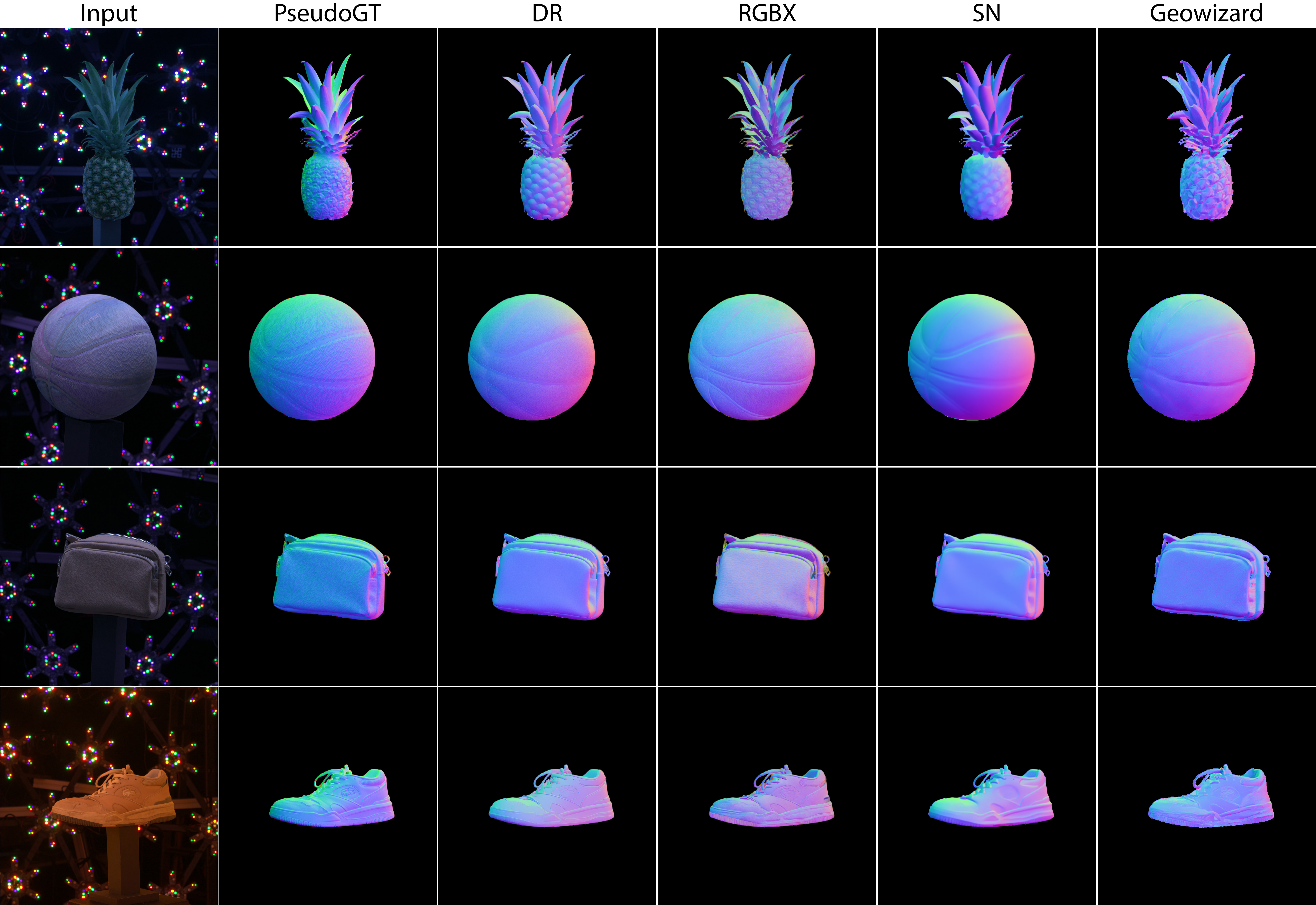} 
    \caption{Visual comparison of pseudo ground truth normals with normals estimated by DR~\cite{liang2025diffusion}, RGBX~\cite{zeng2024rgb}, SN~\cite{ye2024stablenormal} and GW~\cite{fu2024geowizard}. To ensure a  robust and generalized comparison, we provide input images of each validation object under four different illuminations. }
    \label{fig:normal}
    \vspace{-0.2in}
\end{figure*}

\subsection{Relighting}
\label{sec:relighting}

With OLATs, we can produce image-based renderings under any arbitrary illuminations, leveraging the linearity of light transport. Specifically, given a target equirectangular environment map $\tgtenv$, the relit image $\relit$ of a certain object can be obtained as:

\begin{equation}
\relit = \sum_{i=1}^{N_{olat}} \big( \aver(\tgtenv \odot \maskolat) \cdot \olat \big),
\label{eq:diff}
\end{equation}

where $\odot$ denotes pixel-wise multiplication, $\aver$ represents per-channel averaging operator, $N_{olat}$ denotes the number of OLATs. $\olat$ and $\maskolat$ denote the $i$th OLAT image and the corresponding environmental mask. Examples of some relighting results are shown in \refFig{teaser}. Leveraging this property, \ourdata{} can be efficiently scaled to a relighting dataset of real objects under diverse illuminations.

\subsection{Inverse Rendering and View Synthesis}
\label{sec:irnvs}

To construct a comprehensive validation dataset, we carefully select $42$ objects with rich textures spanning $14$ material categories. For each object, the raw images are downsampled to $750 \times 1.4\text{k}$ resolution, and every fifth camera and light are used as inference sets, while the remaining data are utilized for training. Object meshes are reconstructed using the feature-based algorithm provided in MetaShape~\cite{agisoft_metashape}.
We evaluate Mitsuba~\cite{nimier2019mitsuba} (using MetaShape-reconstructed meshes) as well as several 3DGS-based inverse rendering methods~\cite{bi2024gs3, fan2025rng, liu2024bigs} on our dataset. All baseline methods are executed using the source code released by the authors and initialized with the same reconstructed meshes.
We visualize relighting results under inference views and lighting conditions in ~\refFig{IR_img}, and report quantitative metrics (SSIM~\cite{wang2004image}, PSNR, and LPIPS~\cite{zhang2018unreasonable}) in ~\refTab{IR_numerical}. As shown, GS$^3$ consistently outperforms other inverse rendering methods both visually and numerically. Compared to the other methods, GS$^3$ accurately captures the specular reflections on glossy surfaces such as the watermelon, metallic rabbit, and plastic roadblock.

\subsection{Normal Estimation}
\label{sec:normal}

Next, we benchmark several diffusion-based normal estimation methods~\cite{liang2025diffusion, zeng2024rgb, ye2024stablenormal, fu2024geowizard} on our validation dataset.  In \ourdata{}, the extracted surface normals from $5$ polarized views and $35$ non-polarized views are in the world coordinate. During preprocessing, we select normal maps extracted from $5$ polarized views for evaluation and reproject the surface normals from world to camera coordinates using camera parameters, ensuring compatibility with the monocular normal estimation methods. To enable lighting-invariant evaluation, each object is evaluated under four different environmental illuminations. For all experiments, we use official source code, and present both qualitative and quantitative results in ~\refFig{normal} and ~\refTab{normal_comp}. We follow the same evaluation protocol as StableNormal (SN)~\cite{ye2024stablenormal}, reporting the mean and median angular errors, where the lower values indicates better accuracy. We also measure the percentage of pixels with angular error below thresholds of $11.25^\circ$, $22.5^\circ$, and $30^\circ$. 
Quantitatively, SN and GeoWizard (GW), which are specifically designed for normal estimation tasks, outperform RGBX and DiffusionRender(DR), which are originally developed for image/video relighting. Visually RGBX and DR produce the surface normals with finer high-frequency details than SN and GW. However, none of these methods recover accurate surface normals for real-world objects with complex geometric structures in the evaluation dataset, highlighting the importance of \ourdata{} for advancing normal estimation research.

\begin{table}[t]
\centering
\caption{Numerical comparison of inverse rendering baselines on our validation dataset using PSNR, LPIPS, and SSIM metrics.}
\vspace{-0.1in}
\begin{tabular}{l|ccc}
\hline
Method & PSNR $\uparrow$ & LPIPS $\downarrow$ & SSIM $\uparrow$ \\
\hline
Mitsuba+Mshape~\cite{nimier2019mitsuba} & 35.906 & 0.0260 & 0.976 \\
\textit{GS$^{3}$}~\cite{bi2024gs3} & 38.538 & 0.0263 & 0.982 \\
RNG~\cite{fan2025rng} & 32.065 & 0.051 & 0.962 \\
BiGS~\cite{liu2024bigs} & 32.98 & 0.0451 & 0.940 \\
\hline
\end{tabular}
\label{tab:IR_numerical}
\end{table}

\begin{table}[t]
\caption{Numerical comparison of normal estimation methods on our validation dataset using normal angular metrics.}
\vspace{-0.1in}
\centering
\setlength{\tabcolsep}{4pt} 
\begin{tabular}{l|cc|ccc}
\hline
Method & Mean$\downarrow$ & Med$\downarrow$ & $11.25^\circ\uparrow$  &  $22.5^\circ\uparrow$ &  $30^\circ\uparrow$ \\
\hline
SN~\cite{ye2024stablenormal} & \textbf{31.85} & \textbf{30.25} & 8.93  & 34.00  & \textbf{55.40}  \\
RGBX~\cite{zeng2024rgb} & 51.95 & 49.70  & 6.40  & 22.80 & 35.85 \\
DR~\cite{liang2025diffusion} & 34.88 & 33.28  & 8.13  & 31.00 & 50.15 \\
GW~\cite{fu2024geowizard} & 34.42 & 32.03 & \textbf{10.98}  & \textbf{34.10} & 50.05 \\
\hline
\end{tabular}
\vspace{-0.1in}
\label{tab:normal_comp}
\end{table}



\section{Limitation}
\label{sec:future}

Although we apply a linear polarized filter to cancel out the specular reflection during normal extraction, artifacts remain noticeable for objects with glossy materials or low-reflectance texture, consistent with the observation in RNHA~\cite{zhou2023relightable}. These artifacts arise from the weak signal-to-noise ratio of low-reflectance objects and the ambiguity between view-dependent reflection and surface normals. While the extracted surface normals and diffuse albedo are not exact ground truth, they can still provide valuable supervision signals for multi-modal training tasks. In addition, the ground truth meshes are not incorporated in our dataset due to hardware limitations. In the future, it would be interesting to integrate an advanced scanning system to jointly capture both appearance and geometry of real objects.

\section{Conclusion and Discussion}
\label{sec:conclusion}

In this work, we introduce \ourdata{}, the first large-scale real-world dataset, comprising around $\imgnumber$ images of $\objnumber$ real objects with diverse physical sizes and materials under precise lighting control. Compared to the existing object-centric datasets, \ourdata{} offers two substantial advantages: large-scale and high-fidelity appearance under precise lighting control. In addition, our dataset provides auxiliary resources such as surface normals, diffuse albedo, and accurate object masks. We demonstrate that \ourdata{} can be used to construct a relighting resource under arbitrary illuminations and serves as a comprehensive real-world benchmark for inverse rendering, view synthesis, relighting, and normal estimation tasks. A promising avenue for future work is to leverage this dataset to train data-driven generative priors for realistic relighting and appearance modeling. We believe that \ourdata{} will advance research toward bridging the gap between synthetic and real-world data in graphics and vision communities.
{
    \small
    \bibliographystyle{ieeenat_fullname}
    \bibliography{main}
}


\end{document}